\makeatletter\@ifundefined{iflatexml}{\newif\iflatexml\latexmlfalse}{}\makeatother
\iflatexml
\documentclass{article}
\usepackage[table]{xcolor}
\usepackage{graphicx,subcaption,bm,placeins,etoolbox,hyperref}
\usepackage[noabbrev,nameinlink]{cleveref}
\usepackage[round,authoryear]{natbib}
\definecolor{greenaccent}{HTML}{1F7A5C}
\definecolor{greenfg}{HTML}{1B2A26}
\newcommand{\beginappendix}{\appendix\section*{Appendix}}
\else
\documentclass[]{greenpreprint}
\fi
\usepackage{amsmath,amssymb}
\usepackage{booktabs,tabularx,array}
\usepackage{float}
\definecolor{resMuted}{HTML}{596574}
\definecolor{resBlue}{HTML}{1F7A5C}  
\definecolor{resOrange}{HTML}{EE601D}
\definecolor{resGreen}{HTML}{049948}
\definecolor{resTint}{HTML}{E2F2E8}  
\newcolumntype{R}{>{\centering\arraybackslash}X}

\usepackage{siunitx,multirow}

\iflatexml\renewcommand{\rowcolor}[2][]{}\fi
\usepackage{url}
\colorlet{resCite}{greenaccent}
\hypersetup{colorlinks=true,citecolor=resCite,linkcolor=greenaccent,urlcolor=greenaccent}

\title{Recursive Environment Scaling for General Long-Horizon Agents}

\iflatexml
\author{}
\date{}
\else
\author[1,*]{Lingxiao Jing}
\author[2,*]{Zhengbo Jiao}
\author[1]{Yichen Xiang}
\author[1]{Bofan Zhu}
\author[3]{Lingbei Meng}
\author[1]{Yan Ni}
\author[4]{Haorui Chen}
\author[1]{Ran Ran}
\author[1]{Zhirong Shen}
\author[5]{Yuheng Yang}
\author[6]{Chenhang Cui}
\author[1]{Yaosong Du}
\author[1]{Rui Huang}

\affiliation[1]{UESTC}
\affiliation[2]{HKUST}
\affiliation[3]{CUHK (SZ)}
\affiliation[4]{THU}
\affiliation[5]{Westlake}
\affiliation[6]{NUS}
\contribution[*]{Equal Contribution}

\fi

\newcommand{\RESabstract}{
We study environment scaling as a way to increase agent-benchmark difficulty while holding core task objectives fixed. Our framework, Recursive Environment Scaling (RES), transforms existing benchmark environments while preserving their objectives and completion requirements. A proposer changes information access, resource relations, and execution dependencies; execution-grounded checks validate the resulting candidate, and accepted changes are inherited across rounds. This produces related environments for paired evaluation and training-trajectory collection. Experiments on 294 initial tasks from three benchmarks show lower task success for all six evaluation models, with decreases of 57.14--77.14 percentage points on Claw-Eval-Live. With equal trajectory counts and identical optimization settings, fine-tuning on RES trajectories outperforms original-trajectory SFT in all nine model--benchmark comparisons by 7.00--14.61 percentage points. Ablations identify recursive inheritance and dependency composition as important contributors to increased difficulty. Together, these results support environment scaling for both agent evaluation and training.
}

\iflatexml
\else
\abstract{\RESabstract}
\correspondence{\href{mailto:frostlin2005@gmail.com}{\texttt{frostlin2005@gmail.com}}, \href{mailto:huang_rui@std.uestc.edu.cn}{\texttt{huang\_rui@std.uestc.edu.cn}}}
\metadata[Code]{\url{https://github.com/Glimmerxx/recursive-environment-scaling}}

\fi

\begin{document}
\maketitle
\iflatexml
\begin{center}
Lingxiao Jing\textsuperscript{1,*}, Zhengbo Jiao\textsuperscript{2,*}, Yichen Xiang\textsuperscript{1}, Bofan Zhu\textsuperscript{1}, Lingbei Meng\textsuperscript{3}, Yan Ni\textsuperscript{1}, Haorui Chen\textsuperscript{4}, Ran Ran\textsuperscript{1}, Zhirong Shen\textsuperscript{1}, Yuheng Yang\textsuperscript{5}, Chenhang Cui\textsuperscript{6}, Yaosong Du\textsuperscript{1}, Rui Huang\textsuperscript{1}\par
\medskip
{\small
\textsuperscript{1} UESTC; \textsuperscript{2} HKUST; \textsuperscript{3} CUHK (SZ); \textsuperscript{4} THU; \textsuperscript{5} Westlake; \textsuperscript{6} NUS\par
\medskip
\textsuperscript{*}Equal contribution: Lingxiao Jing and Zhengbo Jiao.\par
\textbf{Correspondence:} {\href{mailto:frostlin2005@gmail.com}{\texttt{frostlin2005@gmail.com}}, \href{mailto:huang_rui@std.uestc.edu.cn}{\texttt{huang\_rui@std.uestc.edu.cn}}}
\par\textbf{Code:} {\url{https://github.com/Glimmerxx/recursive-environment-scaling}}

\par}
\end{center}
\section*{Abstract}
\RESabstract
\medskip
\fi

\begin{figure}[H]
\centering
\includegraphics[width=\linewidth]{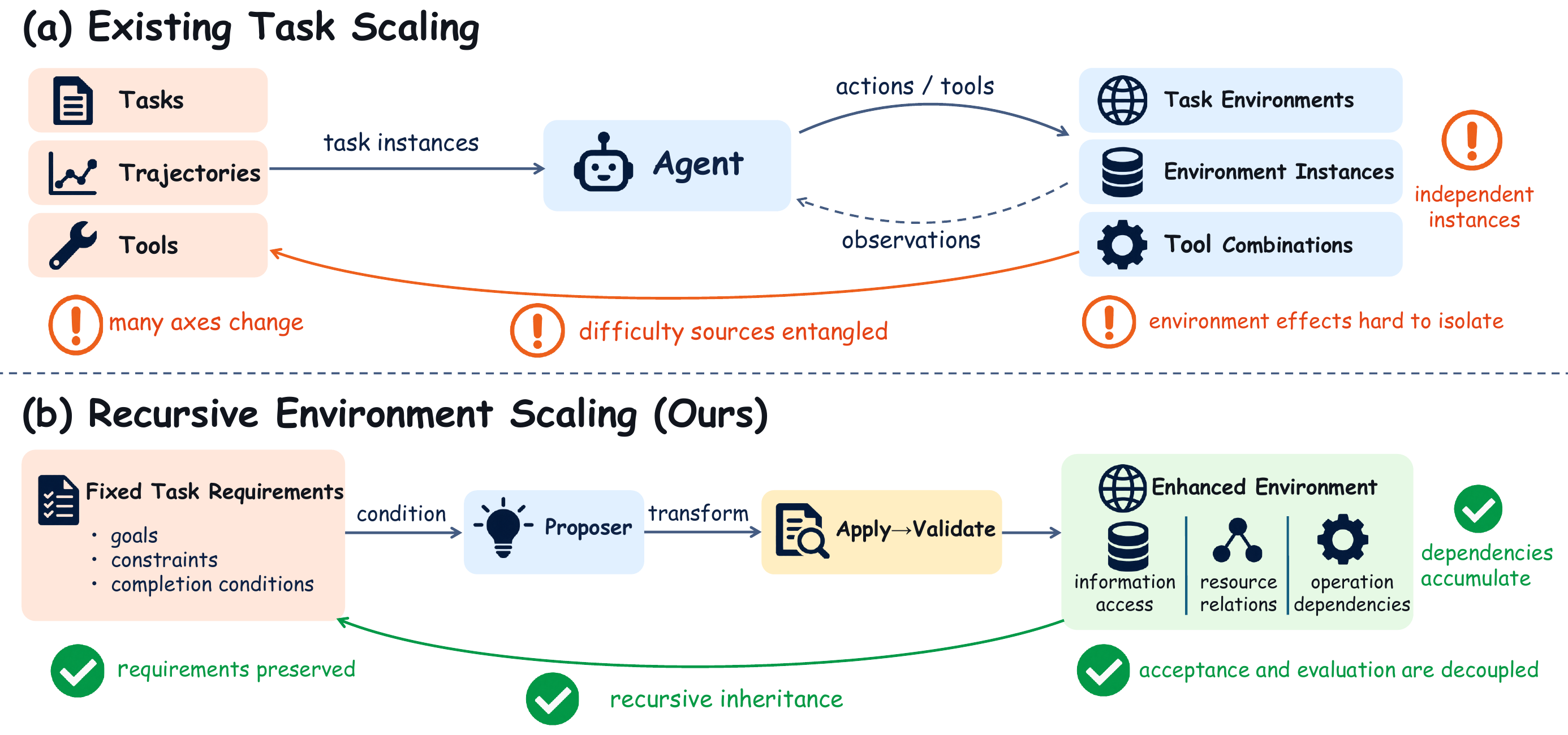}
\caption{\textbf{Task scaling and recursive environment scaling.} Task scaling expands tasks and trajectories. RES preserves core objectives while recursively transforming information access, resource relations, and execution dependencies. Accepted changes accumulate across rounds.}
\label{fig:res-comparison}
\end{figure}

\begin{figure}[t!]
\centering
\includegraphics[width=\linewidth]{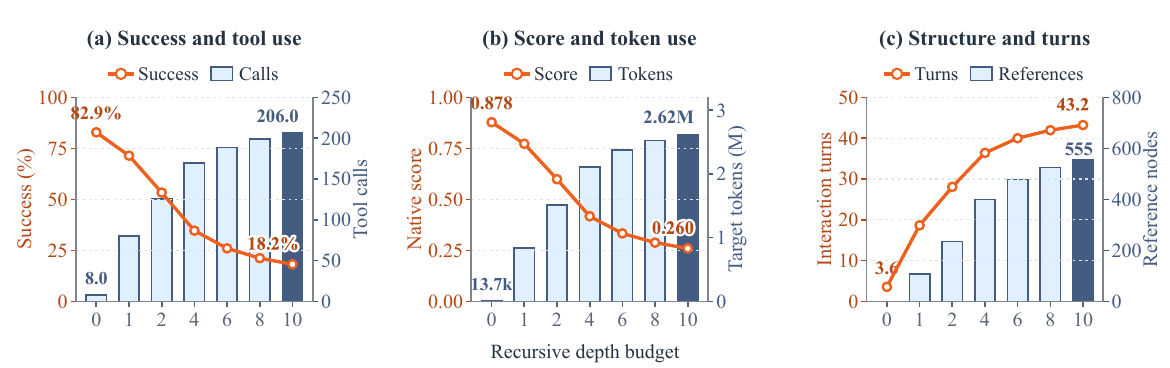}
\caption{\textbf{Environment scaling increases execution demand.} Historical means for 105 Claw-Eval-Live tasks using DeepSeek-V4.1-Flash (one attempt, 50 turns, success threshold 0.80). Budget $d$ averages retained states $0,\ldots,d$, carrying forward early-stopped finals (Appendix~\ref{app:current-states}). Tokens include target input and output; reference nodes count added structure, \mbox{not shortest solution paths.}}
\label{fig:intro-quantitative}
\end{figure}

\section{Introduction}
Language model agents increasingly perform tasks through extended interaction with digital environments. AgentBench~\citep{agentbench} examines decisions across diverse environments, OSWorld~\citep{osworld} tests real computer use, and TheAgentCompany~\citep{theagentcompany} evaluates workplace workflows. Across these settings, difficulty depends on how information is organized, resources are connected, and earlier actions constrain later ones. This makes the environment itself a useful object of study when scaling agent evaluation.

Recent work expands agent experience through task synthesis and environment construction. TaskCraft~\citep{taskcraft} scales the depth and breadth of tool-use tasks; LOGIGEN~\citep{logigen} grounds task construction in executable policies and state verification. RST~\citep{rst} recursively extends reference solutions and realigns instructions and verifiers with the resulting workflows. At the environment level, EnvScaler~\citep{envscaler} synthesizes tool-interaction sandboxes, and ScaleEnv~\citep{scaleenv} constructs interactive environments and verifiable tasks from scratch. RES transforms existing benchmark environments under fixed task objectives and completion requirements, enabling paired evaluation across descendants.

WorkArena++~\citep{workarenapp} composes multi-step knowledge-work tasks, while $\tau^2$-Bench~\citep{tau2bench} couples agent and user actions through shared state. These settings highlight how dependencies shape task execution. Changing access paths and prerequisites can lengthen a task's workflow while preserving the information and operations needed for completion. This motivates our research question: How can we recursively scale executable environments while preserving core task objectives?

We introduce \textbf{Recursive Environment Scaling (RES)} (Figure~\ref{fig:res-comparison}). A proposer inspects a validated parent environment and composes programmatic transformations of its resources and execution dependencies. Execution-grounded checks determine whether the candidate realizes the proposed changes and preserves task validity. Accepted children are inherited for subsequent construction, allowing later transformations to build on earlier structures. Target-agent evaluation measures difficulty separately from environment acceptance. The lineages support paired evaluation and training-trajectory collection. Figure~\ref{fig:intro-quantitative} summarizes performance, interaction demand, and environment structure on Claw-Eval-Live.

Across six models and three benchmarks, RES reduces task success in all 18 comparisons. On Claw-Eval-Live, DeepSeek-V4.1-Flash falls from 82.86\% to 5.71\%. With equal-sized training sets, RES trajectories also improve all three Qwen3.5 Base models over original-trajectory SFT on every benchmark. Our contributions are threefold:
\begin{enumerate}
\item \textbf{Environment Scaling.} We study existing benchmark environments as a scaling dimension, preserving core task objectives while transforming information access and execution dependencies.
\item \textbf{RES Framework.} We combine proposer-guided transformations, execution-grounded validation, and recursive inheritance to accumulate accepted changes along each task lineage.
\item \textbf{Empirical Evidence.} RES reduces task success across six models and improves SFT across three backbones; ablations assess inheritance and dependencies.
\end{enumerate}
\begin{figure}[tb!]
\centering
\includegraphics[width=\linewidth]{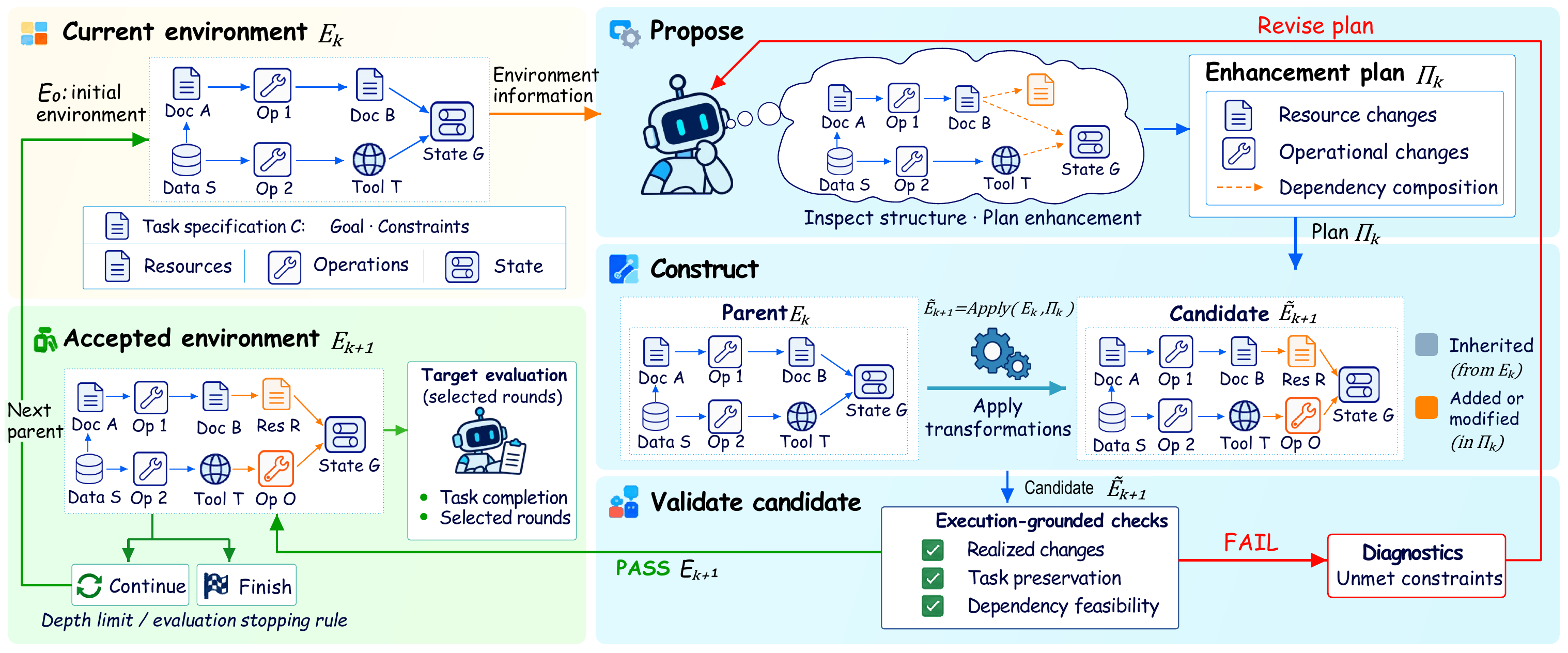}
\caption{\textbf{Overview of RES.} The proposer transforms accepted parent $E_k$. Execution-grounded checks validate the changes, task validity, and dependency feasibility, guiding revisions before accepting a child. Accepted children become parents; target evaluation separately informs stopping.}
\label{fig:res-overview}
\end{figure}

\section{Related Work}

\paragraph{Evaluating agents in executable environments.}
SWE-bench~\citep{swebench} and Terminal-Bench~\citep{terminalbench} cover repository and command-line tasks. WebArena~\citep{webarena} and WebVoyager~\citep{webvoyager} evaluate website interaction, with BrowserGym~\citep{browsergym} providing shared interfaces. WorkArena~\citep{workarena} studies enterprise workflows; $\tau$-Bench~\citep{taubench} couples tool use with domain policies and final-state checks. AppWorld~\citep{appworld} spans applications, while Claw-Eval-Live~\citep{clawevallive} and Workspace-Bench~\citep{workspace} emphasize execution evidence and file dependencies.

\paragraph{Task and trajectory construction.}
Mind2Web~\citep{mind2web} collects human action sequences across websites. TaskCraft~\citep{taskcraft} expands task depth and breadth, while LOGIGEN~\citep{logigen} grounds construction in executable policies and state verification. DIVE~\citep{dive} uses executed tool traces, and CLI-Universe~\citep{cliverse} grounds terminal tasks in technical material. RST~\citep{rst} extends reference solutions recursively and realigns instructions and verifiers. RES keeps core task requirements fixed while transforming the environment.

\paragraph{Environment construction and adaptation.}
SWE-Gym~\citep{swegym} packages tasks with runtimes and tests, and AgentGym~\citep{agentgym} supports exploration and training across environments. EnvScaler~\citep{envscaler} synthesizes tool-interaction sandboxes; ScaleEnv~\citep{scaleenv} builds tools, dependency graphs, and verifiable tasks from scratch. Agent-World~\citep{agentworld} combines environment discovery with a training arena, and Gym-Anything~\citep{gymanything} configures real software for computer use. SPICE~\citep{spice} grounds self-play in a document corpus, while SPADE~\citep{spade} trains an environment designer jointly with a learner. RES recursively inherits accepted changes to existing benchmarks, supporting paired difficulty evaluation and training-trajectory collection.

\FloatBarrier
\section{Method}

\subsection{Problem Formulation and Overview}\label{sec:method-1}
We represent a task at recursion step $k$ by $(C_k,E_k)$, where $C_k$ specifies its objective, constraints, and completion conditions, and $E_k$ contains the executable resources, states, operations, and interaction dependencies. RES fixes the core objective, underlying facts, and completion requirements; specification updates describe the access instructions needed after environmental changes. A validated seed starts a \mbox{recursive environment lineage:}
\begin{equation}
(C_0,E_0)\rightarrow(C_1,E_1)\rightarrow\cdots\rightarrow(C_K,E_K).
\label{eq:task-environment-lineage}
\end{equation}
The proposer composes a transformation plan $\Pi_k$ for the accepted parent. Executing this plan produces a candidate,
\begin{equation}
\widetilde E_{k+1}=\operatorname{Apply}(E_k,\Pi_k),
\label{eq:candidate-overview}
\end{equation}
which becomes the next parent only after validation (Section~\ref{sec:method-3}). Figure~\ref{fig:res-overview} illustrates the procedure.
Public instructions describe revised access paths and prerequisites, allowing the agent to recover the information needed for the original required outputs.

\subsection{Environment Transformation}\label{sec:method-2}
\paragraph{Planning and execution.}
The proposer conditions its plan on the task specification and current environment, selecting programmatic primitives, resource bindings, parameters, and their composition. Resource transformations change information access; relational transformations alter relationships among sources; operational transformations modify when actions become available. The plan binds transformations to concrete resources and orders their execution. Applying it produces the interfaces and state transitions that the target encounters (Equation~\ref{eq:candidate-overview}). Benchmark-specific resource bindings and the transformation library are described in Appendices~\ref{app:protocol} and~\ref{app:operator-library}.

\paragraph{Dependency composition.}
RES connects transformations so that one operation can consume information or state produced by another. Dependencies may combine several prerequisites before enabling an action or propagate state changes to later operations. Subsequent generations can extend these structures, adding interaction requirements while retaining a valid solution path.
For example, an evidence view may require a link query and export preparation. A later transformation can add index recovery before that view. The inherited procedure becomes part of a longer execution path around the same objective, as the reconciliation case in Section~\ref{sec:case-study} illustrates.

\subsection{Validated Environment Construction}\label{sec:method-3}
\paragraph{Task preservation and validation.}
Acceptance checks task preservation, construction consistency, and public executability. The verifier checks the parent lineage and protected task requirements, then independently replays the declared transformations from the fixed parent. The rebuilt executable definition must match the submitted candidate. We write the validation decision as
\begin{equation}
v_k=V(C_{k+1},E_k,\Pi_k,\widetilde E_{k+1};\xi_k),
\label{eq:validation}
\end{equation}
where $\xi_k$ denotes replay and execution evidence. In a fresh session, the verifier follows public interfaces to resolve references, execute preparation steps, and reconstruct transformed resources. Recovered content is compared with source facts; missing records, invalid references, and unsatisfied prerequisites reject the candidate. These checks cover the transformed access paths while preserving the original completion requirements (Appendix~\ref{app:protocol}).

\paragraph{Repair.}
Failed checks return diagnostics about unsatisfied requirements, construction mismatches, or invalid dependencies. The proposer revises the affected resource bindings, parameters, or operation sequence and resubmits the candidate. Repair changes the proposed child while preserving the accepted parent. Candidates still invalid at the repair limit are discarded, leaving that parent available for subsequent construction.

\subsection{Recursive Environment Scaling}\label{sec:method-4}
The validated child becomes the next parent, preserving its resources and dependencies. Later plans reuse existing access paths and attach prerequisites to inherited operations, so accepted changes accumulate along the lineage.

We evaluate the original and last accepted environments, plus intermediate generations requested by the proposer, under the protocol in Appendix~\ref{app:protocol}. At measured nodes, three assessed failures with no measurement errors stop further descendants. Transformation planning and acceptance use the environment and validation evidence. Recursion ends at a budget or stopping condition, with the last accepted snapshot retained as the scaled environment.

\FloatBarrier

\section{Experiments}
\label{sec:experiments}
We test whether RES makes existing benchmarks more demanding across six evaluation models and improves training across three Qwen3.5 Base backbones.

\begin{table}[tb!]
\centering\small
\caption{\textbf{Environment scaling lowers task completion across six models and three benchmarks.} Original denotes the released benchmark environment; +RES denotes the last accepted environment after scaling. Each model makes one 50-turn attempt per task in both conditions. All metrics use a 0--100 scale. Claw Pass@1 requires native score $\geq0.80$, and Score is the mean native score. Workspace TCR@80 requires at least 80\% of a task's rubric items to pass; RPR is the percentage of passed rubric items pooled across tasks. Terminal Pass@1 uses binary verifier outcomes.}
\label{tab:main-results}
\setlength{\tabcolsep}{2.5pt}\renewcommand{\arraystretch}{1.12}
\begin{tabularx}{\linewidth}{@{}l l R R R R R@{}}
\toprule
 & & \multicolumn{2}{c}{\shortstack{\textbf{Claw-Eval-Live}\\$N=105$}} & \multicolumn{2}{c}{\shortstack{\textbf{Workspace-Bench-Lite}\\$N=100$}} & \shortstack{\textbf{Terminal-}\\\textbf{Bench 2.1}\\$N=89$} \\
\cmidrule(lr){3-4}\cmidrule(lr){5-6}\cmidrule(l){7-7}
Model & Environment & Pass@1 & Score & TCR@80 & RPR & Pass@1 \\
\midrule
\mbox{DeepSeek-V4.1-Flash} & Original & 82.86 & 87.71 & 63.00 & 71.19 & 91.01 \\
\rowcolor{resTint} & +RES & 5.71 & 14.20 & 46.00 & 60.38 & 43.82 \\
\addlinespace[2pt]
GPT-5.6 Sol & Original & 85.71 & 91.24 & 61.00 & 70.60 & 88.76 \\
\rowcolor{resTint} & +RES & 26.67 & 35.44 & 49.00 & 65.08 & 44.94 \\
\addlinespace[2pt]
Claude Opus 5 & Original & 86.67 & 92.13 & 70.00 & 74.98 & 89.89 \\
\rowcolor{resTint} & +RES & 28.57 & 37.28 & 51.00 & 65.57 & 47.19 \\
\addlinespace[2pt]
GLM-5.2 & Original & 80.95 & 88.64 & 59.00 & 69.30 & 80.90 \\
\rowcolor{resTint} & +RES & 23.81 & 39.80 & 45.00 & 61.19 & 39.33 \\
\addlinespace[2pt]
Qwen3.8-Max & Original & 82.86 & 90.82 & 67.00 & 72.81 & 86.52 \\
\rowcolor{resTint} & +RES & 22.86 & 33.23 & 49.00 & 63.08 & 41.57 \\
\addlinespace[2pt]
Kimi K3 & Original & 84.76 & 90.16 & 65.00 & 72.38 & 88.76 \\
\rowcolor{resTint} & +RES & 13.33 & 23.85 & 48.00 & 62.70 & 42.70 \\
\bottomrule
\end{tabularx}
\end{table}

\subsection{Experimental Setup}
\label{sec:experimental-setup}

\paragraph{Benchmarks and models.}
We use Claw-Eval-Live~\citep{clawevallive}, Workspace-Bench-Lite~\citep{workspace}, and Terminal-Bench 2.1~\citep{terminalbench}, with 105, 100, and 89 tasks, respectively. Each task defines a paired comparison between its original environment and the last environment accepted by RES. We evaluate DeepSeek-V4.1-Flash, GPT-5.6 Sol, Claude Opus 5, GLM-5.2, Qwen3.8-Max, and Kimi K3 on both versions. Task success measures whether an agent satisfies the benchmark's completion criterion; native scores and rubric pass rates additionally measure partial completion. The table captions define each metric.

\paragraph{Scaling and evaluation protocol.}
GPT-5.6 Sol proposes environment transformations, and DeepSeek-V4.1-Flash serves as the scaling-stage target; both use high reasoning effort. Recursion proceeds to at most ten accepted transformations. Original and last accepted environments are always evaluated; the proposer selects any intermediate evaluations. Three 60-turn target attempts inform stopping during scaling. We then freeze the environments and evaluate each model with one 50-turn attempt per task and high reasoning effort, holding tool permissions and scoring fixed within each pair. Appendix~\ref{app:protocol} gives additional execution settings and validity checks.

\paragraph{Training comparison.}
We fine-tune Qwen3.5-4B-Base, Qwen3.5-9B-Base, and Qwen3.5-35B-A3B-Base under two conditions. \textit{+SFT w/o RES} uses original ClawGym trajectories; \textit{+SFT w/ RES} uses only trajectories collected by DeepSeek-V4.1-Flash in RES-scaled environments. Both start from the same Base checkpoint and use equal numbers of trajectories with identical optimization settings. All checkpoints are evaluated on the original tasks of the three benchmarks with the same single-attempt protocol. Data selection and hyperparameters appear in Appendix~\ref{app:training-settings}.

\subsection{Scaling Benchmark Difficulty}
\label{sec:main-results}

\paragraph{Environment scaling reduces task success across models.}
Table~\ref{tab:main-results} shows lower success in all 18 model--benchmark comparisons. On Claw-Eval-Live, DeepSeek-V4.1-Flash drops from 82.86\% to 5.71\%, while Claude Opus 5 drops from 86.67\% to 28.57\%. The reduction spans 12.00--19.00 percentage points on Workspace-Bench-Lite and 41.57--47.19 on Terminal-Bench 2.1. Under the same single-attempt evaluation budget, the scaled environments are consistently harder across the model suite.

\paragraph{Requirement completion also declines.}
Every model obtains a lower Claw native score and Workspace RPR after scaling. For GPT-5.6 Sol, Claw native score falls from 91.24 to 35.44; Workspace RPR reductions range from 5.52 to 10.81 points across the model suite. Agents therefore satisfy fewer underlying requirements as well as complete fewer tasks. Section~\ref{sec:recursive-trends} examines how this decline develops with recursive depth and increased interaction demand.

\subsection{Training on Scaled Environments}
\label{sec:training-utility}
\begin{table}[tb!]
\centering\small
\caption{\textbf{SFT with RES trajectories improves performance on the original benchmarks.} For each Qwen3.5 Base model, +SFT w/o RES uses original-environment trajectories; +SFT w/ RES uses only trajectories collected in RES-scaled environments. Both use the same initialization, sample count, and optimization. Each checkpoint makes one 50-turn attempt per task. Metrics use a 0--100 scale: Claw Pass@1 thresholds native scores at 0.80, Score averages native scores, Workspace TCR@80 thresholds task rubric completion at 80\%, RPR is the pooled percentage of passed rubric items, and Terminal Pass@1 uses its binary verifier. Bold marks the best result per model.}
\label{tab:sft}
\setlength{\tabcolsep}{2pt}\renewcommand{\arraystretch}{1.12}
\begin{tabularx}{\linewidth}{@{}l *{9}{R}@{}}
\toprule
 & \multicolumn{3}{c}{\textbf{Qwen3.5-4B}} & \multicolumn{3}{c}{\textbf{Qwen3.5-9B}} & \multicolumn{3}{c}{\textbf{Qwen3.5-35B-A3B}} \\
\cmidrule(lr){2-4}\cmidrule(lr){5-7}\cmidrule(l){8-10}
Metric & Base & \shortstack{+SFT\\w/o RES} & \shortstack{+SFT\\w/ RES} & Base & \shortstack{+SFT\\w/o RES} & \shortstack{+SFT\\w/ RES} & Base & \shortstack{+SFT\\w/o RES} & \shortstack{+SFT\\w/ RES} \\
\midrule
\multicolumn{10}{@{}l}{\textit{Claw-Eval-Live ($N=105$)}} \\
Pass@1 & 49.52 & 52.38 & \cellcolor{resTint}\textbf{61.90} & 60.00 & 62.86 & \cellcolor{resTint}\textbf{71.43} & 61.90 & 64.76 & \cellcolor{resTint}\textbf{77.14} \\
Score & 71.51 & 74.12 & \cellcolor{resTint}\textbf{79.15} & 77.75 & 80.32 & \cellcolor{resTint}\textbf{81.04} & 78.96 & 80.36 & \cellcolor{resTint}\textbf{82.83} \\
\addlinespace[3pt]
\multicolumn{10}{@{}l}{\textit{Workspace-Bench-Lite ($N=100$)}} \\
TCR@80 & 1.00 & 4.00 & \cellcolor{resTint}\textbf{11.00} & 9.00 & 12.00 & \cellcolor{resTint}\textbf{21.00} & 11.00 & 14.00 & \cellcolor{resTint}\textbf{28.00} \\
RPR & 7.46 & 26.59 & \cellcolor{resTint}\textbf{33.68} & 30.97 & 35.89 & \cellcolor{resTint}\textbf{42.22} & 34.70 & 37.24 & \cellcolor{resTint}\textbf{45.62} \\
\addlinespace[3pt]
\multicolumn{10}{@{}l}{\textit{Terminal-Bench 2.1 ($N=89$)}} \\
Pass@1 & 2.25 & 5.62 & \cellcolor{resTint}\textbf{13.48} & 10.11 & 14.61 & \cellcolor{resTint}\textbf{24.72} & 13.48 & 23.60 & \cellcolor{resTint}\textbf{38.20} \\
\bottomrule
\end{tabularx}
\end{table}

\paragraph{RES trajectories improve SFT across all three backbones.}
Training with RES trajectories outperforms original-trajectory SFT in all nine model--benchmark comparisons (Table~\ref{tab:sft}). Averaged over the three backbones, the gains are 10.16, 10.00, and 10.86 percentage points on Claw-Eval-Live, Workspace-Bench-Lite, and Terminal-Bench 2.1. The largest gains occur for 35B-A3B: 12.38, 14.00, and 14.61 points, respectively.

\paragraph{Training gains extend to partial completion.}
Claw native score and Workspace RPR improve for every backbone. For 4B, Workspace RPR rises from 26.59 to 33.68 while TCR@80 rises from 4.00 to 11.00. The gains therefore span both rubric-level completion and the number of tasks meeting the success threshold.

\FloatBarrier
\section{Ablations and Analysis}
\label{sec:analysis}
\subsection{Ablation Studies}
\label{sec:ablations}
We examine dependency composition, recursive inheritance, depth, and library breadth on Claw-Eval-Live (Table~\ref{tab:component-ablations}). DeepSeek-V4.1-Flash makes one attempt per environment with a 50-turn limit and success threshold 0.80. Component comparisons share 15 tasks; depth analysis uses 105 frozen lineages. Protocols and per-task results appear in Appendices~\ref{app:ablation-protocol} and~\ref{app:per-task-results}.

\begin{table}[tb!]
\centering
\caption{\textbf{Effects of RES components and recursion depth on Claw-Eval-Live.} DeepSeek-V4.1-Flash makes one attempt per environment with a 50-turn limit. Pass@1 uses native score $\geq0.80$; Score is mean native score on a 0--100 scale. Lower values indicate harder environments. A removes composed dependencies and their prerequisite work from saved environments. B uses the same 15 tasks with matched planned depths and proposer budgets: w/o inheritance evaluates the last independent enhancement from the original environment, and Basic retains six operators. Checkmarks mark enabled components; blue shading identifies full RES controls and the default depth cap of ten. C reevaluates 105 separate lineages, retaining the last accepted environment after early stopping.}
\label{tab:component-ablations}
\small
\setlength{\tabcolsep}{4pt}
\begingroup
\renewcommand{\arraystretch}{1.13}
\begin{tabularx}{\linewidth}{@{}X c c c r r@{}}
\toprule
Variant & \shortstack{Dependency\\composition} & Inheritance & Library & \shortstack{Pass@1 (\%)\\$\downarrow$} & \shortstack{Score\\$\downarrow$} \\
\midrule
\multicolumn{6}{@{}l}{\textbf{A. Remove dependencies from saved environments}\quad(15 tasks)} \\
\rowcolor{resTint}Saved RES & $\checkmark$ & $\checkmark$ & Full & \textbf{0.00} & \textbf{7.73} \\
w/o dependencies & $\times$ & $\checkmark$ & Full & 100.00 & 92.93 \\
\addlinespace[4pt]
\multicolumn{6}{@{}l}{\textbf{B. Rebuild environments under matched budgets}\quad(15 tasks)} \\
\rowcolor{resTint}Rebuilt RES & $\checkmark$ & $\checkmark$ & Full & \textbf{6.67} & \textbf{14.40} \\
w/o inheritance & $\checkmark$ & $\times$ & Full & 93.33 & 88.73 \\
w/ basic library & $\checkmark$ & $\checkmark$ & Basic & 80.00 & 85.73 \\
\addlinespace[4pt]
\multicolumn{6}{@{}l}{\textbf{C. Increase the maximum recursion depth}\quad(105 tasks)} \\
\multicolumn{4}{@{}l}{Original benchmark (depth 0)} & 82.86 & 87.84 \\
\multicolumn{4}{@{}l}{Maximum depth 1} & 60.00 & 66.69 \\
\multicolumn{4}{@{}l}{Maximum depth 2} & 17.14 & 25.17 \\
\multicolumn{4}{@{}l}{Maximum depth 3} & 7.62 & 15.30 \\
\multicolumn{4}{@{}l}{Maximum depth 5} & 3.81 & 11.84 \\
\rowcolor{resTint}\multicolumn{4}{@{}l}{Maximum depth 10} & 4.76 & 13.37 \\
\bottomrule
\end{tabularx}
\endgroup

\end{table}

\paragraph{Dependency composition increases difficulty.}
Removing cross-operator references and preparation prerequisites from saved RES environments, while retaining task facts and atomic transformation histories, raises pass@1 from 0.0\% to 100.0\% and mean native score from 7.73 to 92.93. Mean turns fall from 50.0 to 19.1, linking the removed prerequisites to both completion and interaction cost.

\paragraph{Recursive inheritance accumulates enhancement.}
For planned depth $d_i$, we compare recursive construction with $d_i$ independent enhancements of the original environment, evaluating the predeclared last construction. Both use the same nominal proposer-call limit per enhancement. The recursive control yields 6.7\% pass@1 versus 93.3\% independently, with scores of 14.40 versus 88.73 and approximately 2.5 times the mean target-token expenditure. Inheriting earlier changes therefore produces harder environments under matched construction opportunities.

\paragraph{Additional depth strengthens enhancement.}
Applying depth caps to the same 105 lineages, with early-stopped finals retained, reduces pass@1 from 82.9\% at $E_0$ to 60.0\% after one round, 17.1\% at depth two, and 4.8\% at depth ten. Mean native score falls from 87.84 to 13.37. These retained-environment measurements show gains beyond one round; Appendix~\ref{app:depth-details} reports all caps and distinguishes them from historical averages.

\paragraph{A richer library enables stronger enhancement.}
Restricting transformations to a Basic subset of six mechanisms, with recursion and planned depths unchanged, raises pass@1 from 6.7\% to 80.0\% and native score from 14.40 to 85.73 relative to the shared regenerated control. The full library gives the proposer more effective ways to compose environmental requirements.
\FloatBarrier

\subsection{Cumulative Effects of Recursive Environment Scaling}
\label{sec:recursive-trends}

\begin{figure}[tb!]
\centering
\includegraphics[width=\linewidth]{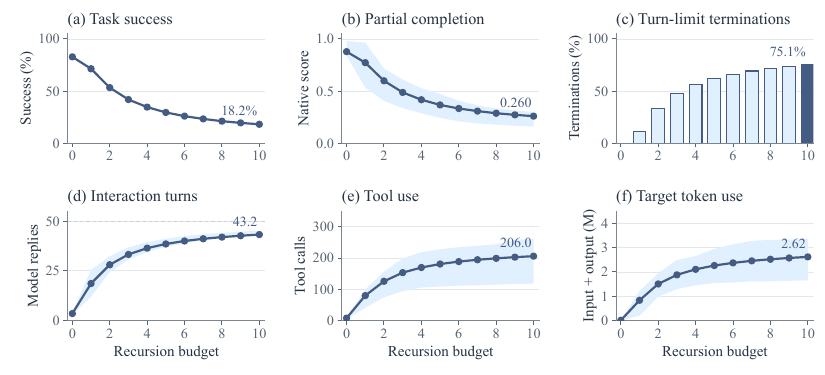}
\caption{\textbf{Historical performance and interaction demand on Claw-Eval-Live.} Lines and bars show 105-seed historical means (Equation~\ref{eq:historical-mean}); shading shows task interquartile ranges. The dashed line marks the 50-turn limit. Tokens include target input and output. Termination rates use logged outcomes; actual depth counts appear in Appendix~\ref{app:interaction}.}
\label{fig:recursive-historical}
\end{figure}

We fix DeepSeek-V4.1-Flash and a 50-turn budget per attempt. Figure~\ref{fig:recursive-historical} reports historical means from $E_0$ through each depth budget over the same 105 tasks, retaining final environments for early-stopped chains (Appendix~\ref{app:current-states}).

\paragraph{Success and partial completion decline together.}
From budget zero to ten, historical mean success falls from 82.86\% to 18.18\%, and mean native score from 0.878 to 0.260. The two measures show a decline in both task success and partial completion under this benchmark-level aggregation.

\paragraph{Interaction demand increases.}
Under the same aggregation, mean turns rise from 3.55 to 43.22 and tool calls from 8.05 to 206.02; the turn-limit rate reaches 75.06\%. These measurements include successful and unsuccessful trajectories. Depth and endpoint evaluations use separate batches; Appendix~\ref{app:current-states} reports the retained-state results and depth counts.

\subsection{Environment Scaling Efficiency and Cost}
\label{sec:construction-cost}
\begin{figure}[tb!]
\centering
\includegraphics[width=\linewidth]{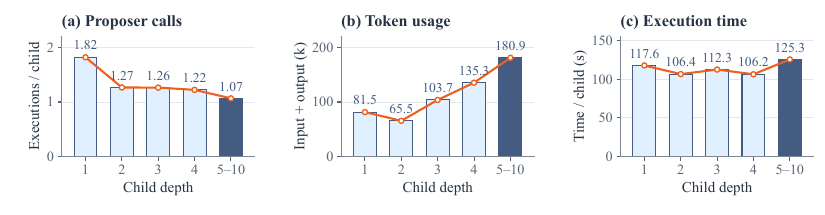}
\caption{\textbf{Proposer cost across recursive stages.} Bars and lines show the same per-child means on Claw-Eval-Live. Depths 1--4 contain 105, 94, 57, and 18 children; dark bars pool 15 children from five seeds at depths 5--10. Time sums proposer execution durations.}
\label{fig:construction-cost}
\end{figure}

On Claw-Eval-Live, RES produces 289 accepted children from 105 seeds. Of these, 92.0\% require at most two outer proposer executions; mean costs are 1.45 executions, 89.2k tokens, and 112.6 seconds per child. Figure~\ref{fig:construction-cost} shows fewer outer executions but greater token use at later stages, so execution counts alone do not capture scaling cost. Adaptive stopping also changes which tasks reach later stages.

Costs cover executions associated with accepted children, excluding unsuccessful branches, target solving, and grading. An outer execution may contain multiple model requests and tool calls; time sums execution durations. Full definitions and distributions appear in Appendix~\ref{app:protocol}.

\subsection{Case Study: Three-Source Financial Reconciliation}
\label{sec:case-study}
\begin{figure}[tb!]
\centering
\includegraphics[width=\linewidth]{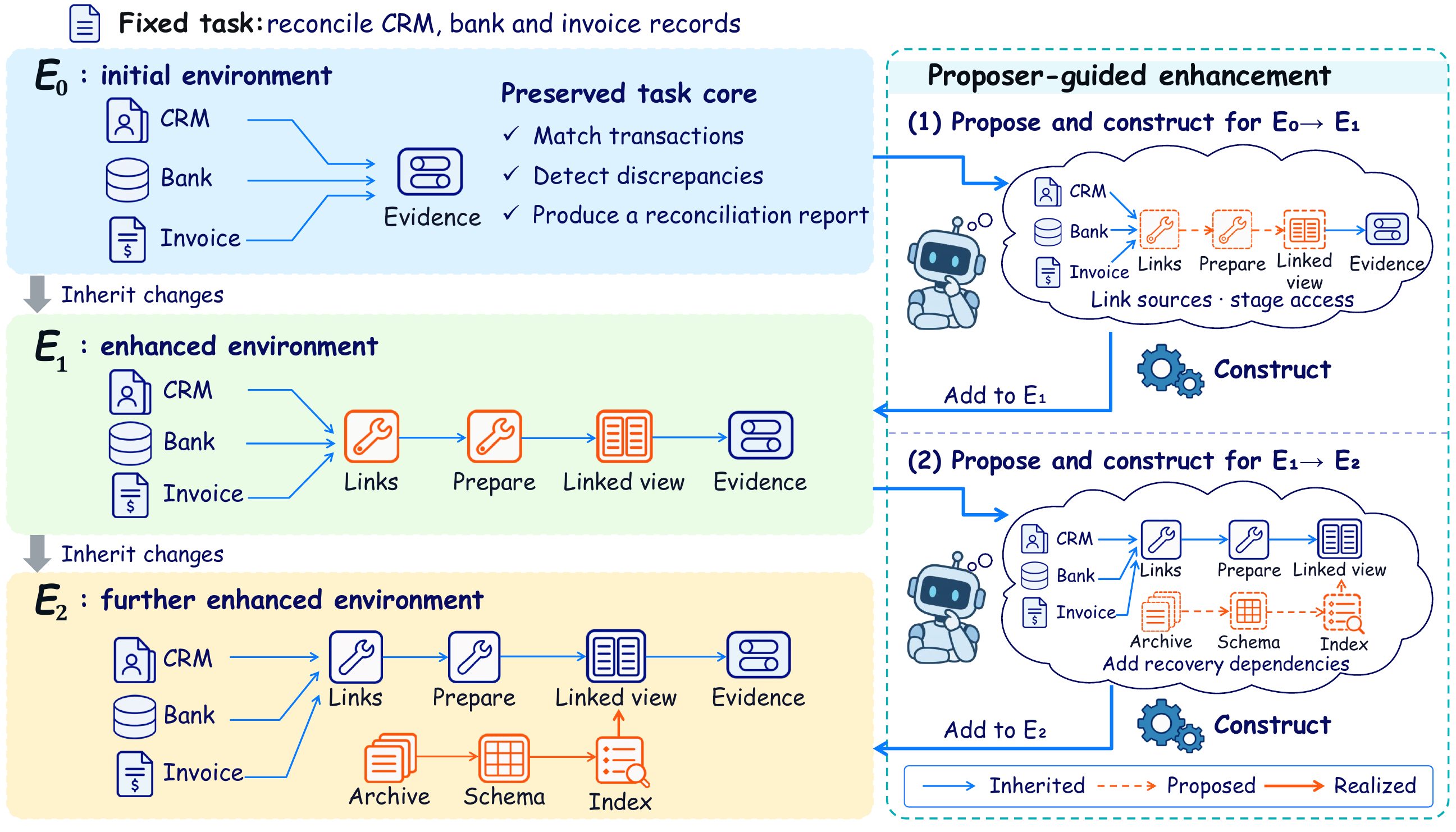}
\caption{\textbf{Recursive scaling of a financial reconciliation task.} Access and recovery dependencies accumulate while the reconciliation objective is preserved. The diagram shows $E_0$--$E_2$; endpoint evaluation uses final $E_3$.}
\label{fig:case-a01}
\end{figure}

Task A01 in Claw-Eval-Live requires matching CRM, bank, and invoice records, identifying discrepancies, and producing a reconciliation report. In $E_0$, the agent can read three CSV files directly. $E_1$ introduces link queries and export preparation before a linked view can be accessed. $E_2$ inherits these steps and adds archive, schema, and index-recovery dependencies (Figure~\ref{fig:case-a01}); final $E_3$ adds encoding constraints. Evidence-access requirements accumulate around the \mbox{preserved core reconciliation objective.}

In the independent paired evaluation, DeepSeek-V4.1-Flash makes 171 tool calls in the scaled environment, compared with three in the original environment, while its native score falls from 0.98 to zero. The trajectory includes reference queries, export preparation, and index rebuilding, with one batched query covering 37 references. The agent reaches the 50-turn limit while still acquiring evidence and does not complete the reconciliation report. The added work is concentrated before reconciliation: the original three reads expose the needed records directly, whereas the scaled environment requires the agent to resolve inherited access and recovery dependencies. This case links the observed increase in interaction demand to the operations introduced along the lineage.

Validation and target evaluation provide complementary evidence in this example. Restoration checks verify that the required records remain recoverable through public operations. The rollout then measures whether the agent can complete the original report within its budget. Its failure during evidence acquisition connects the lower score to the added access procedure. The objective and required deliverable stay fixed throughout this comparison.

\section{Conclusion}
\label{sec:conclusion}
RES makes existing executable environments a controllable dimension of agent evaluation and training. Under preserved core task objectives, recursively inherited transformations reduce task completion across three benchmarks and six models, while the resulting training trajectories improve all three Qwen3.5 Base models over original-trajectory SFT. Component ablations and trajectory analysis connect these observations to inherited dependencies and increased interaction demand. These results support environment scaling as a practical way to extend existing benchmarks and collect training trajectories that improve agent performance.

\subsection*{AI Use Statement}
Generative AI tools were used to assist with migrating the manuscript's LaTeX source to the preprint template, improving language and presentation, checking internal consistency, and verifying bibliographic metadata and the resulting build. The authors reviewed all suggested changes, made the scientific and editorial decisions, and take responsibility for the final content.

\bibliographystyle{iclr2027_conference}
\bibliography{references}

\begin{thebibliography}{29}
\providecommand{\natexlab}[1]{#1}
\providecommand{\url}[1]{\texttt{#1}}
\expandafter\ifx\csname urlstyle\endcsname\relax
  \providecommand{\doi}[1]{doi: #1}\else
  \providecommand{\doi}{doi: \begingroup \urlstyle{rm}\Url}\fi

\bibitem[Aggarwal et~al.(2026)Aggarwal, Neubig, and Welleck]{gymanything}
Pranjal Aggarwal, Graham Neubig, and Sean Welleck.
\newblock {Gym-Anything: Turn any Software into an Agent Environment}.
\newblock \emph{\href{https://arxiv.org/abs/2604.06126}{arXiv:~2604.06126}}, 2026.

\bibitem[Barres et~al.(2025)Barres, Dong, Ray, Si, and Narasimhan]{tau2bench}
Victor Barres, Honghua Dong, Soham Ray, Xujie Si, and Karthik Narasimhan.
\newblock {$\tau^2$-Bench: Evaluating Conversational Agents in a Dual-Control Environment}.
\newblock \emph{\href{https://arxiv.org/abs/2506.07982}{arXiv:~2506.07982}}, 2025.

\bibitem[Boisvert et~al.(2024)Boisvert, Thakkar, Gasse, Caccia, De~Chezelles, Cappart, Chapados, Lacoste, and Drouin]{workarenapp}
L\'{e}o Boisvert, Megh Thakkar, Maxime Gasse, Massimo Caccia, Thibault Le~Sellier De~Chezelles, Quentin Cappart, Nicolas Chapados, Alexandre Lacoste, and Alexandre Drouin.
\newblock {WorkArena++: Towards Compositional Planning and Reasoning-based Common Knowledge Work Tasks}.
\newblock In \emph{\href{https://proceedings.neurips.cc/paper_files/paper/2024/hash/0b82662b6c32e887bb252a74d8cb2d5e-Abstract-Datasets_and_Benchmarks_Track.html}{Advances in Neural Information Processing Systems}}, 2024.

\bibitem[Chen et~al.(2026)Chen, Zhang, Liu, Chen, Du, Li, Zhong, Wang, Zhu, Song, et~al.]{dive}
Aili Chen, Chi Zhang, Junteng Liu, Jiangjie Chen, Chengyu Du, Yunji Li, Ming Zhong, Qin Wang, Zhengmao Zhu, Jiayuan Song, et~al.
\newblock {DIVE: Scaling Diversity in Agentic Task Synthesis for Generalizable Tool Use}.
\newblock \emph{\href{https://arxiv.org/abs/2603.11076}{arXiv:~2603.11076}}, 2026.

\bibitem[de~Chezelles et~al.(2025)de~Chezelles, Gasse, Lacoste, Caccia, Drouin, Boisvert, Thakkar, Marty, Assouel, Shayegan, Jang, L{\`u}, Yoran, Kong, Xu, Reddy, Neubig, Cappart, Salakhutdinov, and Chapados]{browsergym}
Thibault Le~Sellier de~Chezelles, Maxime Gasse, Alexandre Lacoste, Massimo Caccia, Alexandre Drouin, L{\'e}o Boisvert, Megh Thakkar, Tom Marty, Rim Assouel, Sahar~Omidi Shayegan, Lawrence~Keunho Jang, Xing~Han L{\`u}, Ori Yoran, Dehan Kong, Frank~F. Xu, Siva Reddy, Graham Neubig, Quentin Cappart, Russ Salakhutdinov, and Nicolas Chapados.
\newblock {The BrowserGym Ecosystem for Web Agent Research}.
\newblock \emph{\href{https://openreview.net/forum?id=5298fKGmv3}{Transactions on Machine Learning Research}}, 2025.

\bibitem[Deng et~al.(2023)Deng, Gu, Zheng, Chen, Stevens, Wang, Sun, and Su]{mind2web}
Xiang Deng, Yu~Gu, Boyuan Zheng, Shijie Chen, Sam Stevens, Boshi Wang, Huan Sun, and Yu~Su.
\newblock {Mind2Web: Towards a Generalist Agent for the Web}.
\newblock In \emph{\href{https://proceedings.neurips.cc/paper_files/paper/2023/hash/5950bf290a1570ea401bf98882128160-Abstract-Datasets_and_Benchmarks.html}{Advances in Neural Information Processing Systems}}, 2023.

\bibitem[Dong et~al.(2026)Dong, Lu, Huang, Zhong, Liu, Huang, Li, Zhao, Song, Li, et~al.]{agentworld}
Guanting Dong, Junting Lu, Junjie Huang, Wanjun Zhong, Longxiang Liu, Shijue Huang, Zhenyu Li, Yang Zhao, Xiaoshuai Song, Xiaoxi Li, et~al.
\newblock {Agent-World: Scaling Real-World Environment Synthesis for Evolving General Agent Intelligence}.
\newblock \emph{\href{https://arxiv.org/abs/2604.18292}{arXiv:~2604.18292}}, 2026.

\bibitem[Drouin et~al.(2024)Drouin, Gasse, Caccia, Laradji, Del~Verme, Marty, Vazquez, Chapados, and Lacoste]{workarena}
Alexandre Drouin, Maxime Gasse, Massimo Caccia, Issam~H. Laradji, Manuel Del~Verme, Tom Marty, David Vazquez, Nicolas Chapados, and Alexandre Lacoste.
\newblock {{W}ork{A}rena: How Capable are Web Agents at Solving Common Knowledge Work Tasks?}
\newblock In \emph{\href{https://proceedings.mlr.press/v235/drouin24a.html}{Proceedings of the 41st International Conference on Machine Learning}}, pp.\  11642--11662, 2024.

\bibitem[He et~al.(2024)He, Yao, Ma, Yu, Dai, Zhang, Lan, and Yu]{webvoyager}
Hongliang He, Wenlin Yao, Kaixin Ma, Wenhao Yu, Yong Dai, Hongming Zhang, Zhenzhong Lan, and Dong Yu.
\newblock {{W}eb{V}oyager: Building an End-to-End Web Agent with Large Multimodal Models}.
\newblock In \emph{\href{https://aclanthology.org/2024.acl-long.371/}{Proceedings of the 62nd Annual Meeting of the Association for Computational Linguistics (Volume 1: Long Papers)}}, 2024.

\bibitem[Hua et~al.(2026)Hua, Yao, Xie, Zhao, Liu, Qiu, Huang, Wang, Ji, Ye, et~al.]{cliverse}
Zhanbo Hua, Yifan Yao, Weihao Xie, Yongchi Zhao, Minghao Liu, Ruizhi Qiu, Zhewei Huang, Zun Wang, Yiyan Ji, Yunhai Ye, et~al.
\newblock {CLI-Universe: Towards Verifiable Task Synthesis Engine for Terminal Agents}.
\newblock \emph{\href{https://arxiv.org/abs/2606.22883}{arXiv:~2606.22883}}, 2026.

\bibitem[Jimenez et~al.(2024)Jimenez, Yang, Wettig, Yao, Pei, Press, and Narasimhan]{swebench}
Carlos~E Jimenez, John Yang, Alexander Wettig, Shunyu Yao, Kexin Pei, Ofir Press, and Karthik Narasimhan.
\newblock {SWE-bench: Can Language Models Resolve Real-world Github Issues?}
\newblock In \emph{\href{https://proceedings.iclr.cc/paper_files/paper/2024/hash/edac78c3e300629acfe6cbe9ca88fb84-Abstract-Conference.html}{International Conference on \mbox{Learning Representations}}}, 2024.

\bibitem[Li et~al.(2026{\natexlab{a}})Li, Tang, Huang, Lin, Huang, Liu, Ye, Li, Li, Wang, and Yuan]{clawevallive}
Chenxin Li, Zhengyang Tang, Mingxin Huang, Yunlong Lin, Shijue Huang, Shengyuan Liu, Bowen Ye, Rang Li, Lei Li, Benyou Wang, and Yixuan Yuan.
\newblock {Claw-Eval-Live: A Live Agent Benchmark for Evolving Real-World Workflows}.
\newblock \emph{\href{https://arxiv.org/abs/2604.28139}{arXiv:~2604.28139}}, 2026{\natexlab{a}}.

\bibitem[Li et~al.(2026{\natexlab{b}})Li, Shi, Li, Wang, Li, Huang, Yang, Ke, Liu, Mi, and Liang]{rst}
Zhongzhi Li, Yucheng Shi, Zongxia Li, Ruhan Wang, Anhao Li, Zixun Huang, Junyao Yang, Lei Ke, Ninghao Liu, Haitao Mi, and Leowei Liang.
\newblock {Recursive Synthesis for Long-Horizon Terminal Tasks}.
\newblock \emph{\href{https://arxiv.org/abs/2608.05466}{arXiv:~2608.05466}}, 2026{\natexlab{b}}.

\bibitem[Liu et~al.(2025)Liu, Jin, Kim, Yuan, Zhao, Kulikov, Li, Sukhbaatar, Lanchantin, and Weston]{spice}
Bo~Liu, Chuanyang Jin, Seungone Kim, Weizhe Yuan, Wenting Zhao, Ilia Kulikov, Xian Li, Sainbayar Sukhbaatar, Jack Lanchantin, and Jason Weston.
\newblock {SPICE: Self-Play In Corpus Environments Improves Reasoning}.
\newblock \emph{\href{https://arxiv.org/abs/2510.24684}{arXiv:~2510.24684}}, 2025.

\bibitem[Liu et~al.(2026)Liu, Yu, Jiang, Qu, Zhao, Liu, Kim, Zhou, Kim, Ren, et~al.]{spade}
Bo~Liu, Simon Yu, Yiding Jiang, Ao~Qu, Andrew Zhao, Zichen Liu, Junsu Kim, Zijian Zhou, Seungone Kim, Tongzheng Ren, et~al.
\newblock {SPADE: Self-Play in Adaptive Synthetic Executable Environments}.
\newblock \emph{\href{https://arxiv.org/abs/2608.19197}{arXiv:~2608.19197}}, 2026.

\bibitem[Liu et~al.(2024)Liu, Yu, Zhang, Xu, Lei, Lai, Gu, Ding, Men, Yang, Zhang, Deng, Zeng, Du, Zhang, Shen, Zhang, Su, Sun, Huang, Dong, and Tang]{agentbench}
Xiao Liu, Hao Yu, Hanchen Zhang, Yifan Xu, Xuanyu Lei, Hanyu Lai, Yu~Gu, Hangliang Ding, Kaiwen Men, Kejuan Yang, Shudan Zhang, Xiang Deng, Aohan Zeng, Zhengxiao Du, Chenhui Zhang, Sheng Shen, Tianjun Zhang, Yu~Su, Huan Sun, Minlie Huang, Yuxiao Dong, and Jie Tang.
\newblock {AgentBench: Evaluating LLMs as Agents}.
\newblock In \emph{\href{https://proceedings.iclr.cc/paper_files/paper/2024/hash/e9df36b21ff4ee211a8b71ee8b7e9f57-Abstract-Conference.html}{International Conference on \mbox{Learning Representations}}}, 2024.

\bibitem[Merrill et~al.(2026)Merrill, Shaw, Carlini, Li, Raj, Bercovich, Shi, Shin, Walshe, Buchanan, et~al.]{terminalbench}
Mike~A. Merrill, Alexander~G. Shaw, Nicholas Carlini, Boxuan Li, Harsh Raj, Ivan Bercovich, Lin Shi, Jeong~Yeon Shin, Thomas Walshe, E.~Kelly Buchanan, et~al.
\newblock {Terminal-Bench: Benchmarking Agents on Hard, Realistic Tasks in Command Line Interfaces}.
\newblock \emph{\href{https://arxiv.org/abs/2601.11868}{arXiv:~2601.11868}}, 2026.

\bibitem[Pan et~al.(2025)Pan, Wang, Neubig, Jaitly, Ji, Suhr, and Zhang]{swegym}
Jiayi Pan, Xingyao Wang, Graham Neubig, Navdeep Jaitly, Heng Ji, Alane Suhr, and Yizhe Zhang.
\newblock {Training Software Engineering Agents and Verifiers with {SWE}-Gym}.
\newblock In \emph{\href{https://proceedings.mlr.press/v267/pan25g.html}{Proceedings of the 42nd International Conference on Machine Learning}}, pp.\  47717--47737, 2025.

\bibitem[Shi et~al.(2025)Shi, Cao, Chen, Sun, Li, Lu, Dong, Qin, Zhu, Liu, et~al.]{taskcraft}
Dingfeng Shi, Jingyi Cao, Qianben Chen, Weichen Sun, Weizhen Li, Hongxuan Lu, Fangchen Dong, Tianrui Qin, King Zhu, Minghao Liu, et~al.
\newblock {TaskCraft: Automated Generation of Agentic Tasks}.
\newblock \emph{\href{https://arxiv.org/abs/2506.10055}{arXiv:~2506.10055}}, 2025.

\bibitem[Song et~al.(2026)Song, Chang, Dong, Zhu, Wen, and Dou]{envscaler}
Xiaoshuai Song, Haofei Chang, Guanting Dong, Yutao Zhu, Ji-Rong Wen, and Zhicheng Dou.
\newblock {EnvScaler: Scaling Tool-Interactive Environments for LLM Agent via Programmatic Synthesis}.
\newblock \emph{\href{https://arxiv.org/abs/2601.05808}{arXiv:~2601.05808}}, 2026.

\bibitem[Tang et~al.(2026)Tang, Zhou, Liu, Li, Wu, Wang, Huang, Zhou, Zhou, Song, et~al.]{workspace}
Zirui Tang, Xuanhe Zhou, Yumou Liu, Linchun Li, Yukai Wu, Weizheng Wang, Hongzhang Huang, Wei Zhou, Jun Zhou, Jiachen Song, et~al.
\newblock {Workspace-Bench 1.0: Benchmarking AI Agents on Workspace Tasks with Large-Scale File Dependencies}.
\newblock \emph{\href{https://arxiv.org/abs/2605.03596}{arXiv:~2605.03596}}, 2026.

\bibitem[Trivedi et~al.(2024)Trivedi, Khot, Hartmann, Manku, Dong, Li, Gupta, Sabharwal, and Balasubramanian]{appworld}
Harsh Trivedi, Tushar Khot, Mareike Hartmann, Ruskin Manku, Vinty Dong, Edward Li, Shashank Gupta, Ashish Sabharwal, and Niranjan Balasubramanian.
\newblock {App{W}orld: A Controllable World of Apps and People for Benchmarking Interactive Coding Agents}.
\newblock In \emph{\href{https://aclanthology.org/2024.acl-long.850/}{ACL}}, 2024.

\bibitem[Tu et~al.(2026)Tu, Hao, Yang, Chen, Zhang, Xia, Yang, Sun, Liu, Shen, et~al.]{scaleenv}
Dunwei Tu, Hongyan Hao, Hansi Yang, Yihao Chen, Yi-Kai Zhang, Zhikang Xia, Yu~Yang, Yueqing Sun, Xingchen Liu, Furao Shen, et~al.
\newblock {ScaleEnv: Scaling Environment Synthesis from Scratch for Generalist Interactive Tool-Use Agent Training}.
\newblock \emph{\href{https://arxiv.org/abs/2602.06820}{arXiv:~2602.06820}}, 2026.

\bibitem[Xi et~al.(2025)Xi, Ding, Chen, Hong, Guo, Wang, Guo, Yang, Liao, He, Gao, Chen, Zheng, Zou, Gui, Zhang, Qiu, Huang, Wu, and Jiang]{agentgym}
Zhiheng Xi, Yiwen Ding, Wenxiang Chen, Boyang Hong, Honglin Guo, Junzhe Wang, Xin Guo, Dingwen Yang, Chenyang Liao, Wei He, Songyang Gao, Lu~Chen, Rui Zheng, Yicheng Zou, Tao Gui, Qi~Zhang, Xipeng Qiu, Xuanjing Huang, Zuxuan Wu, and Yu-Gang Jiang.
\newblock {{A}gent{G}ym: Evaluating and Training Large Language Model-based Agents across Diverse Environments}.
\newblock In \emph{\href{https://aclanthology.org/2025.acl-long.1355/}{Proceedings of the 63rd Annual Meeting of the Association for Computational Linguistics (Volume 1: Long Papers)}}, 2025.

\bibitem[Xie et~al.(2024)Xie, Zhang, Chen, Li, Zhao, Cao, Hua, Cheng, Shin, Lei, Liu, Xu, Zhou, Savarese, Xiong, Zhong, and Yu]{osworld}
Tianbao Xie, Danyang Zhang, Jixuan Chen, Xiaochuan Li, Siheng Zhao, Ruisheng Cao, Toh~Jing Hua, Zhoujun Cheng, Dongchan Shin, Fangyu Lei, Yitao Liu, Yiheng Xu, Shuyan Zhou, Silvio Savarese, Caiming Xiong, Victor Zhong, and Tao Yu.
\newblock {OSWorld: Benchmarking Multimodal Agents for Open-Ended Tasks in Real Computer Environments}.
\newblock In \emph{\href{https://proceedings.neurips.cc/paper_files/paper/2024/hash/5d413e48f84dc61244b6be550f1cd8f5-Abstract-Datasets_and_Benchmarks_Track.html}{Advances in Neural Information Processing Systems}}, 2024.

\bibitem[Xu et~al.(2025)Xu, Song, Li, Tang, Jain, Bao, Wang, Zhou, Guo, Cao, Yang, Lu, Martin, Su, Maben, Mehta, Chi, Jang, Xie, Zhou, and Neubig]{theagentcompany}
Frank~(Fangzheng) Xu, Yufan Song, Boxuan Li, Yuxuan Tang, Kritanjali Jain, Mengxue Bao, Zora Wang, Xuhui Zhou, Zhitong Guo, Murong Cao, Mingyang Yang, Hao~Yang Lu, Amaad Martin, Zhe Su, Leander Maben, Raj Mehta, Wayne Chi, Lawrence Jang, Yiqing Xie, Shuyan Zhou, and Graham Neubig.
\newblock {TheAgentCompany: Benchmarking LLM Agents on Consequential Real World Tasks}.
\newblock In \emph{\href{https://proceedings.neurips.cc/paper_files/paper/2025/hash/0d744742f6fac4d1134c019b7cef3c8a-Abstract-Datasets_and_Benchmarks_Track.html}{Advances in Neural Information Processing Systems}}, 2025.

\bibitem[Yao et~al.(2024)Yao, Shinn, Razavi, and Narasimhan]{taubench}
Shunyu Yao, Noah Shinn, Pedram Razavi, and Karthik Narasimhan.
\newblock {$\tau$-bench: A Benchmark for Tool-Agent-User Interaction in Real-World Domains}.
\newblock \emph{\href{https://arxiv.org/abs/2406.12045}{arXiv:~2406.12045}}, 2024.

\bibitem[Zeng et~al.(2026)Zeng, Lu, Liu, Li, Qu, Zhu, Li, Tan, Liu, Zhao, Zhou, and Wu]{logigen}
Yucheng Zeng, Weipeng Lu, Linyun Liu, Shupeng Li, Zitian Qu, Chenghao Zhu, Shaofei Li, Zhengdong Tan, Mengyue Liu, Haotian Zhao, Zhe Zhou, and Jianmin Wu.
\newblock {LOGIGEN: Logic-Driven Generation of Verifiable Agentic Tasks}.
\newblock \emph{\href{https://arxiv.org/abs/2603.00540}{arXiv:~2603.00540}}, 2026.

\bibitem[Zhou et~al.(2024)Zhou, Xu, Zhu, Zhou, Lo, Sridhar, Cheng, Ou, Bisk, Fried, Alon, and Neubig]{webarena}
Shuyan Zhou, Frank~F Xu, Hao Zhu, Xuhui Zhou, Robert Lo, Abishek Sridhar, Xianyi Cheng, Tianyue Ou, Yonatan Bisk, Daniel Fried, Uri Alon, and Graham Neubig.
\newblock {WebArena: A Realistic Web Environment for Building Autonomous Agents}.
\newblock In \emph{\href{https://proceedings.iclr.cc/paper_files/paper/2024/hash/4410c0711e9154a7a2d26f9b3816d1ef-Abstract-Conference.html}{International Conference on \mbox{Learning Representations}}}, 2024.

\end{thebibliography}
\clearpage
\beginappendix
\suppressfloats[t]
\setcounter{figure}{0}
\renewcommand{\thefigure}{A\arabic{figure}}
\setcounter{table}{0}
\renewcommand{\thetable}{A\arabic{table}}
\renewcommand{\theHfigure}{appendix.\arabic{figure}}
\renewcommand{\theHtable}{appendix.\arabic{table}}
\section{Evaluation and Environment Scaling}
\label{app:protocol}

\paragraph{Evaluation protocol.}
We separate the target evaluations used to stop environment scaling from the independent evaluations used to report benchmark performance. During scaling, three attempts test complete task fulfilment. After scaling, we freeze the environments and run one attempt per task. On Claw-Eval-Live, independent evaluation counts native scores of at least 0.80 as successful. Table~\ref{tab:protocol} summarizes the scaling, depth-reevaluation, and model-comparison settings.

\begin{table}[htb]
\centering\small
\caption{\textbf{Protocols for environment scaling and independent evaluation on Claw-Eval-Live.} Scaling-stage attempts determine when to stop recursion; depth reevaluation measures changes over saved lineages; the paired model comparison evaluates original and RES-scaled environments. A model turn is one target-model response, potentially containing multiple tool calls. Stopping uses complete fulfilment; independent success uses a native-score threshold of 0.80.}
\label{tab:protocol}
\setlength{\tabcolsep}{4pt}\renewcommand{\arraystretch}{1.18}
\begin{tabularx}{\linewidth}{@{}>{\raggedright\arraybackslash}p{0.225\linewidth}*{3}{>{\raggedright\arraybackslash}X}@{}}
\toprule
 & Environment scaling & Depth reevaluation & Model comparison \\
\midrule
Attempts per environment & 3 at evaluated nodes & 1 & 1 \\
Model-turn limit & 60 & 50 & 50 \\
Success criterion & Complete task fulfilment & Native score $\geq0.80$ & Native score $\geq0.80$ \\
Target & DeepSeek-V4.1-Flash & DeepSeek-V4.1-Flash & Six evaluation models \\
Task cohort & 105 seeds & 105 seeds / 394 environments & 105 matched seeds per model \\
Primary use & Stopping decisions & Recursive trends & Original vs. RES \\
\bottomrule
\end{tabularx}
\end{table}

\paragraph{Scaling configuration.}
GPT-5.6 Sol with high reasoning effort proposes transformations. Each lineage has a maximum depth of 10 and a budget of eight outer proposer executions per parent, including repairs and retries. We process 32 seeds concurrently; the three target attempts within each seed run sequentially. Every lineage includes evaluation of its original and last accepted environments, with intermediate evaluations requested by the proposer. Target outcomes determine stopping; the validation procedure in Section~\ref{sec:method-3} determines whether a child is accepted.

\paragraph{Target configuration.}
Across all three benchmarks, target attempts have a 60-turn limit during scaling and a 50-turn limit during independent evaluation, with at most 16,384 output tokens per response. Timeouts follow the task-specific Claw configuration: 180, 240, 300, 420, or 600 seconds, held fixed within each original/scaled pair. All six evaluation models use high reasoning effort. DeepSeek-V4.1-Flash uses thinking with high effort throughout scaling, depth reevaluation, and the frozen endpoint comparison.

\paragraph{Workspace and terminal evaluation.}
Workspace-Bench-Lite contains 100 tasks and Terminal-Bench 2.1 contains 89. The latter uses the official 2.1 revision, which fixes 28 tasks from the 2.0 release.\footnote{\href{https://www.tbench.ai/news/terminal-bench-2-1}{Terminal-Bench 2.1 release notes}.} Both follow the scaling and target configurations above. Paired environments use the same model settings, tool permissions, and scoring protocol. Workspace TCR@80 counts tasks that pass at least 80\% of their rubric items; RPR pools passed items across tasks. Terminal-Bench uses its executable verifier's binary result. Success rates weight tasks equally.

\paragraph{Validation and stopping.}
Claw validation checks lineage and protected-contract hashes, replays the declared operators from the frozen parent, and compares the rebuilt files with the candidate. The original goal, facts, and grader remain fixed. Public traversal reconstructs files and service records, rejecting content loss or duplication, invalid references, nonterminating preparation, and original-file bypasses. The record stores replay hashes and execution traces. Stopping requires three assessed failures with no measurement errors.

\paragraph{Workspace and terminal adapters.}
Workspace transformations act on files, structured records, and workflow dependencies. The adapter binds derived views to source hashes and checks restored files or projected records against frozen inputs; Office and PDF resources are inspected with native readers. Reported TCR@80 and RPR use the frozen benchmark rubric. Terminal transformations act on container files, build configurations, database states, and service inputs. Frozen parent state and declared resource bindings identify the inputs to each child. Materialized artifacts undergo native readback and execution checks, and task scoring uses the executable verifier. Private tests and reference solutions remain on the evaluator side.

\paragraph{Environment validity and manual review.}
Three researchers reviewed 20 matched original/scaled pairs from each benchmark: 60 task lineages and 120 endpoint environments in total. They checked core objectives, required information, scoring conditions, answer leakage, and scoring bypasses, alongside programmatic checks of executability and task--evaluation consistency. None of these issues was found in the reviewed samples.

\begin{table}[tb!]
\centering\small
\caption{\textbf{Proposer construction cost per accepted Claw-Eval-Live child.} Mean, median, and 90th-percentile (P90) costs cover all 289 children constructed by GPT-5.6 Sol with high reasoning effort. Each child's costs aggregate its outer executions and retries. Tokens include input and output; time sums execution durations. Target solving and grading are excluded.}
\label{tab:construction-stats}
\renewcommand{\arraystretch}{1.18}
\begin{tabular*}{\linewidth}{@{\extracolsep{\fill}}lrrr@{}}
\toprule
Quantity & Mean & Median & P90 \\
\midrule
Outer proposer executions & 1.45 & 1.00 & 2.00 \\
Input + output tokens (k) & 89.19 & 65.27 & 179.82 \\
Summed execution time (s) & 112.61 & 93.56 & 184.00 \\
\bottomrule
\end{tabular*}
\end{table}

\paragraph{Construction cost.}
The 289 accepted children account for 420 outer executions, 25.78M input and output tokens, and 9.04 hours of summed proposer execution time. We first add the executions associated with each child, including retries, then compute the statistics in Table~\ref{tab:construction-stats}. Of these children, 193 require one execution and 73 require two, accounting for 92.0\% together. Input totals include cached tokens; reasoning tokens are included once in the reported totals. Time measures summed execution durations under concurrent processing. These costs cover accepted-child construction and exclude branches producing no child, target solving, grading, and orchestration overhead.

\section{Training Data and Optimization}
\label{app:training-settings}
\paragraph{Training conditions.}
At each model scale, SFT w/o RES and SFT w/ RES start independently from the same Base checkpoint and use 8,428 trajectories with identical optimization settings (Table~\ref{tab:sft-config}). The original trajectories are drawn from ClawGym and generated by DeepSeek-V4.1-Flash. For SFT w/ RES, RES is applied to this training task pool and DeepSeek-V4.1-Flash collects new trajectories in the scaled environments, using the same selection and supervision rules. Evaluation uses the original environments of the 294 benchmark tasks in Table~\ref{tab:sft}.

\paragraph{Original-trajectory data.}
We select 8,428 distinct tasks from the released ClawGym task and trajectory collections, retaining one structurally complete rollout per task.\footnote{Frozen ClawGym releases: \href{https://huggingface.co/datasets/RUC-AIBOX/ClawGym-Task/tree/bd9043cff88b913922a165d9c022a4796aee3f3a}{tasks (\texttt{bd9043cff88b})} and \href{https://huggingface.co/datasets/RUC-AIBOX/ClawGym-Trajectory/tree/31824a4516a3d06630fee03ff53180fc5fc31287}{trajectories (\texttt{31824a4516a3})}.}
We deduplicate records across release formats and require a traceable task association, complete messages, matched tool calls and returns, and a released native score of at least 0.8. Filtering uses the released scores without replaying the trajectories. The selected data contain 3,381 full-score and 5,047 partial-score trajectories, each no longer than 49,152 tokens. This original-trajectory set contains 190.52M input tokens and 39.74M supervised target tokens; its median sequence length is 20,195 tokens.

\paragraph{Supervision.}
Training preserves the available reasoning, tool calls, tool outputs, and final responses. Loss applies to assistant reasoning, responses, and unmasked tool calls. System and user messages and tool outputs provide context. Calls with missing parameters or unverified semantics remain in context but are excluded from the loss; the original-trajectory set contains 78 such calls from 54 tasks. A custom ms-swift template preserves these token-level supervision labels.

\paragraph{Optimization.}
We run one epoch of full-parameter SFT in BF16 on four H800 GPUs. A per-device batch of one with four accumulation steps gives a global batch of 16, with 12 examples in the last of 527 updates. We evaluate the final checkpoint; training uses no held-out validation split. Table~\ref{tab:sft-config} lists the hyperparameters. The software stack consists of ms-swift 4.6.0.dev0, PyTorch 2.11.0, Transformers 5.16.1, and DeepSpeed 0.19.6.

\begin{table}[tb!]
\centering\small
\caption{\textbf{Optimization settings for the Qwen3.5 Base checkpoints in the training comparison.} +SFT w/o RES and +SFT w/ RES start independently from the same checkpoint and use these identical settings at each model size. Each condition trains on 8,428 trajectories for one epoch; sequence lengths and effective batch size are measured in tokens and examples, respectively. The checkpoint after the single training epoch is used for evaluation.}
\label{tab:sft-config}
\setlength{\tabcolsep}{4pt}\renewcommand{\arraystretch}{1.1}
\begin{tabularx}{\linewidth}{@{}l *{3}{>{\centering\arraybackslash}X}@{}}
\toprule
Setting & 4B-Base & 9B-Base & 35B-A3B-Base \\
\midrule
Peak learning rate & $2\times10^{-5}$ & $10^{-5}$ & $5\times10^{-6}$ \\
Epoch / updates & \multicolumn{3}{c}{1 / 527} \\
Maximum sequence length & \multicolumn{3}{c}{49,152 tokens} \\
Effective batch size & \multicolumn{3}{c}{16 (12 in the last update)} \\
Learning-rate schedule & \multicolumn{3}{c}{Cosine, 3.00\% warmup (16 steps)} \\
Optimizer & \multicolumn{3}{c}{DeepSpeedCPUAdam (AdamW)} \\
Adam $\beta_1,\beta_2,\epsilon$ & \multicolumn{3}{c}{$0.90,\ 0.95,\ 10^{-8}$} \\
Weight decay / gradient clip & \multicolumn{3}{c}{0.01 / 1.00} \\
Precision / hardware & \multicolumn{3}{c}{BF16 / $4\times$ H800} \\
Memory and attention & \multicolumn{3}{c}{ZeRO-3, CPU optimizer offload, FlashAttention 2} \\
Gradient checkpointing & \multicolumn{3}{c}{Enabled (non-reentrant)} \\
Packing / truncation & \multicolumn{3}{c}{Disabled / overlength records excluded} \\
Training / data seed & \multicolumn{3}{c}{42 / 42} \\
\bottomrule
\end{tabularx}
\end{table}

\paragraph{Evaluation-overlap checks.}
We screened the training tasks against all three evaluation benchmarks and found no matches under the checks performed: 105 Claw-Eval-Live, 100 English Workspace-Bench-Lite, and 89 Terminal-Bench 2.1 tasks. The Claw and Workspace records compare task identifiers, task text, input-content hashes, and lexical similarity. These checks address the recorded task and input representations.

\section{Recursive Evaluation and Aggregation}
\label{app:current-states}
\subsection{Current and Historical Statistics}

We reevaluate 394 distinct environments from 105 Claw-Eval-Live seeds using DeepSeek-V4.1-Flash, with one 50-turn attempt per environment. For each seed $i$, let $m_i(r)$ be its measurement at depth $r$ and $D_i$ its final depth. At budget $d$, a stopped lineage retains its last accepted environment. The current-state mean is
\begin{equation}
M_m(d)=\frac{1}{N}\sum_{i=1}^{N}m_i\!\left(\min(d,D_i)\right),\qquad N=105.
\label{eq:retained-mean}
\end{equation}
Figure~\ref{fig:recursive-historical} uses the historical mean, which averages each task's retained measurements over budgets zero through $d$ and then averages equally across tasks:
\begin{equation}
H_m(d)=\frac{1}{N}\sum_{i=1}^{N}\frac{1}{d+1}
\sum_{r=0}^{d}m_i\!\left(\min(r,D_i)\right).
\label{eq:historical-mean}
\end{equation}
At budget two, each task averages its measurements at budgets zero, one, and two. All six panels follow this rule, using binary values for success and turn-limit termination. Stopped lineages reuse their final measurement. Token and interaction statistics are retained-state averages; bands show the interquartile range across task-level historical means.

\begin{figure}[htb!]
\centering
\includegraphics[width=\linewidth]{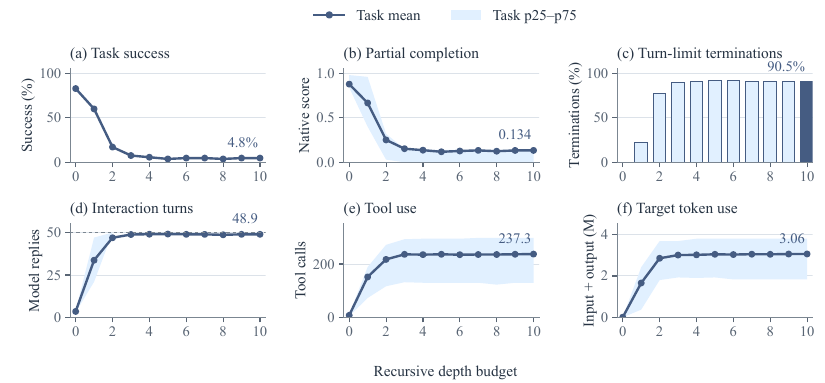}
\caption{\textbf{Current retained-state performance and interaction demand.} Means and interquartile ranges cover all 105 seeds, retaining early-stopped finals without averaging earlier budgets.}
\label{fig:current-retained}
\end{figure}

\begin{figure}[htb!]
\centering
\includegraphics[width=0.96\linewidth]{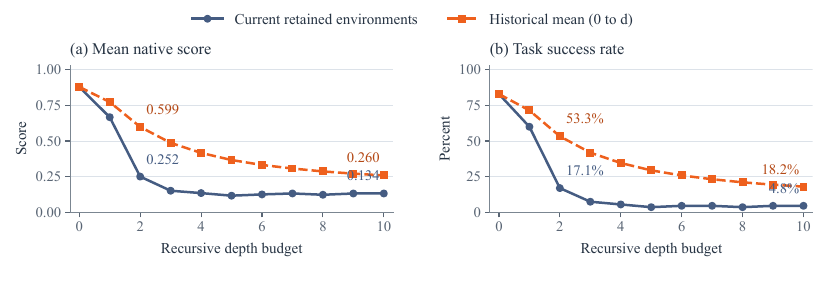}
\caption{\textbf{Current versus historical benchmark means.} Historical means average the retained measurements from budgets zero through the current budget.}
\label{fig:current-history-comparison}
\end{figure}

Both statistics keep all 105 seeds: $M_m$ measures the retained environment and $H_m$ includes earlier states. Figures~\ref{fig:current-retained} and~\ref{fig:current-history-comparison} compare these views. At budget ten, this depth batch has 4.76\% current-state success; the separate endpoint batch in Table~\ref{tab:main-results} gives 5.71\%.

\subsection{Interaction Demand at Selected Budgets}
\label{app:interaction}

\begin{figure}[htb!]
\centering
\includegraphics[width=\linewidth]{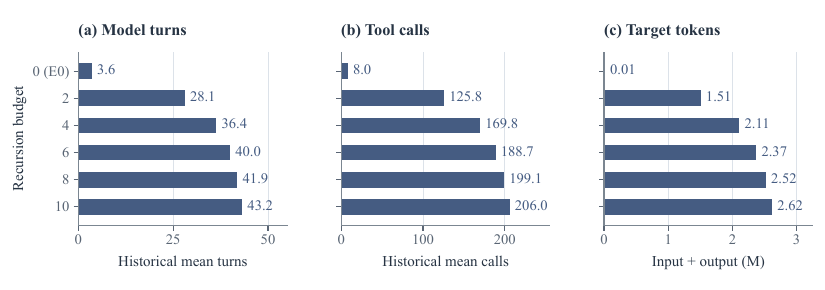}
\caption{\textbf{Historical interaction demand at selected budgets.} Per-task means over retained states $0,\ldots,d$ are averaged equally across 105 seeds, including failed and turn-limited trajectories. Tokens include repeated input context.}
\label{fig:interaction-budgets}
\end{figure}

Figure~\ref{fig:interaction-budgets} compares turns, tool calls, and tokens at budgets 0, 2, 4, 6, 8, and 10 using the historical aggregation in Equation~\ref{eq:historical-mean}.

Among the 105 lineages, 11 stop at depth one, 37 at two, 39 at three, 13 at four, one at five, two at six, one at eight, and one at ten. The numbers reaching depths zero through ten are $105,105,94,57,18,5,4,2,2,1,1$. Higher-budget summaries therefore combine the final environments of many stopped lineages with the few that continue. Keeping all seeds gives a benchmark-level view as the allowed recursion budget increases.

\subsection{Performance Across Depth Budgets}
\label{app:depth-details}

\begin{figure}[htb!]
\centering
\includegraphics[width=\linewidth]{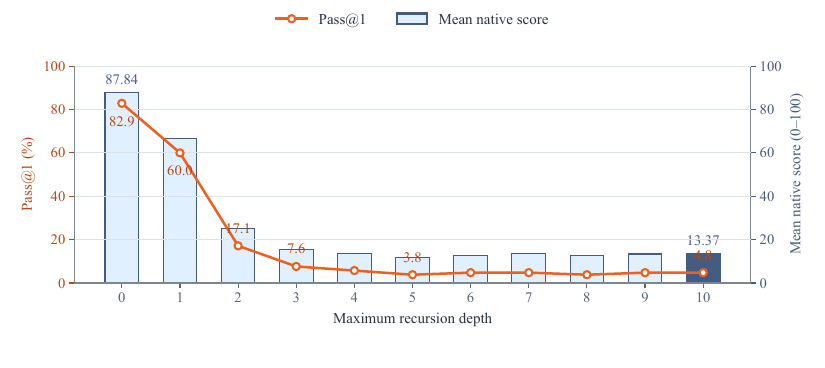}
\caption{\textbf{Performance across recursion-depth caps.} Pass@1 (orange) and mean native score (blue) use a 0--100 scale over 105 seeds, retaining stopped environments. Table~\ref{tab:component-ablations} selects caps 0, 1, 2, 3, 5, and 10.}
\label{fig:depth-ablation}
\end{figure}

We apply the current-state statistic in Equation~\ref{eq:retained-mean} to the same 105 saved lineages. For pass@1, $m_i(r)=\mathbb{1}[s_i(r)\geq0.80]$; for mean native score, $m_i(r)=s_i(r)$. Both use the final depth $D_i$ and budget $d$ defined above, retaining the last accepted environment after stopping. Table~\ref{tab:all-depth-caps} reports all budgets, and Figure~\ref{fig:depth-ablation} plots the resulting performance.

\begin{table}[tb!]
\centering\small
\setlength{\tabcolsep}{4pt}
\caption{\textbf{Performance and interaction cost across recursion-depth budgets on Claw-Eval-Live.} DeepSeek-V4.1-Flash receives one 50-turn attempt per environment. Each budget $d$ covers 105 tasks, retaining stopped environments. Reach $d$ counts chains attaining depth $d$; Continue counts chains exceeding it. Pass@1 counts scores $\geq0.80$; Score is the native mean on a 0--100 scale. Turns, Tools, and Tokens are task means; tokens sum target input and output in millions.}
\label{tab:all-depth-caps}
\begin{tabular*}{\linewidth}{@{\extracolsep{\fill}}rrrrrrrr@{}}
\toprule
$d$ & Reach $d$ & Continue & Pass@1 & Score & Turns & Tools & Tokens (M) \\
\midrule
0 & 105 & 105 & 87/105 & 87.84 & 3.55 & 8.05 & 0.01 \\
1 & 105 & 94 & 63/105 & 66.69 & 33.71 & 151.79 & 1.66 \\
2 & 94 & 57 & 18/105 & 25.17 & 46.91 & 217.47 & 2.85 \\
3 & 57 & 18 & 8/105 & 15.30 & 48.81 & 236.37 & 3.00 \\
4 & 18 & 5 & 6/105 & 13.65 & 49.00 & 235.34 & 3.02 \\
5 & 5 & 4 & 4/105 & 11.84 & 49.10 & 236.54 & 3.04 \\
6 & 4 & 2 & 5/105 & 12.71 & 48.91 & 235.10 & 3.03 \\
7 & 2 & 2 & 5/105 & 13.40 & 48.90 & 235.77 & 3.05 \\
8 & 2 & 1 & 4/105 & 12.48 & 48.65 & 235.73 & 3.04 \\
9 & 1 & 1 & 5/105 & 13.37 & 48.92 & 236.75 & 3.05 \\
10 & 1 & 0 & 5/105 & 13.37 & 48.92 & 237.30 & 3.06 \\
\bottomrule
\end{tabular*}

\end{table}

\paragraph{Evaluation batches.}
At cap ten, this batch gives 5/105 passes and mean score 0.134, versus 6/105 and 0.142 in the separate endpoint-comparison batch (Table~\ref{tab:main-results}). Each batch independently evaluates the original environments using the single-attempt, 50-turn protocol in Appendix~\ref{app:protocol}.

\section{Component Ablation Details}
\label{app:ablation-protocol}
\subsection{Settings and Interventions}

\paragraph{Task selection.}
We use 15 complete cases from 20 frozen lineages, selected by endpoint availability and matching planned depths after evaluation, following the original task-group order. Selection does not use target scores. A01 and D01 lack a matching Basic endpoint, C02 requires supplemental independent construction, and RESEARCH\_01 and W03 fall outside the 15-task quota. The subset contains 2 file/multi-table, 3 cross-service, 4 finance/personnel, 3 planning, and 3 operations/workspace tasks. Table~\ref{tab:per-task-scores} lists the selected cases and planned depths.

\paragraph{Evaluation and generation budgets.}
DeepSeek-V4.1-Flash evaluates each component condition with thinking and high reasoning effort, a 50-turn limit, and at most 16,384 output tokens per response. It also supplies the native grade; a weighted score of at least 0.80 counts as a pass. GPT-5.6 Sol proposes transformations with high reasoning effort and a nominal budget of eight outer Codex executions per parent, covering queries, construction, repairs, and retries. We record proposer tokens without imposing a token cap, so matching construction opportunities does not imply equal realized token cost.

The depth analysis is a separate batch of 394 task-environment evaluations with reported outcomes. It uses DeepSeek-V4.1-Flash with thinking and high reasoning effort, a 50-turn limit, and a 16,384-token response cap, following Appendix~\ref{app:protocol}.

\paragraph{Removing dependencies.}
We resolve cross-operator references into directly usable inputs and release cross-resource preparation prerequisites in saved RES environments. Task instructions, underlying facts, and atomic transformation histories are retained. Model-free restoration checks verify environment consistency before target evaluation. This intervention removes both the composed access dependencies and the work needed to satisfy their prerequisites.

\paragraph{Removing recursive inheritance.}
For task $i$ with planned depth $d_i$, we compare
\begin{align*}
\text{Recursive:}\quad &E_0\rightarrow E_1\rightarrow\cdots\rightarrow E_{d_i},\\
\text{Independent:}\quad &E_1^{(j)}=\operatorname{Enhance}(E_0;j),\quad j=1,\ldots,d_i.
\end{align*}
We evaluate $E_{d_i}$ for the recursive arm and the predeclared last construction $E_1^{(d_i)}$ for the independent arm. Each independent attempt starts from the original environment, and construction order fixes the evaluated endpoint. Planned depths sum to 66 across the 15 tasks, giving both arms 66 construction opportunities with the same nominal per-parent execution budget. Neither arm uses target feedback during construction. The inheritance and library comparisons share a rebuilt RES control; dependency removal uses the saved RES control.

\paragraph{Transformation library.}\label{app:operator-library}
Full contains 17 predefined Claw operators. Basic retains six: sharding, pagination, summary-to-detail access, field references, numeric re-encoding, and export preparation. The additional eleven provide evidence dispersion, namespace bridges, source routing, candidate intersection, dependent lookups, multi-source exports, external-schema views, archive views, version selection, index reconstruction, and staged release. The proposer selects applicable operators from resource capabilities and composes them through explicit source bindings and prerequisite relations. The library is fixed during construction. Basic and Full use the same runtime, verifier, proposer settings, and planned task depths; their comparison includes both operator availability and the proposer's choices among supported compositions.

\subsection{Per-task Results}
\label{app:per-task-results}

\begin{figure}[htb!]
\centering
\includegraphics[width=\linewidth]{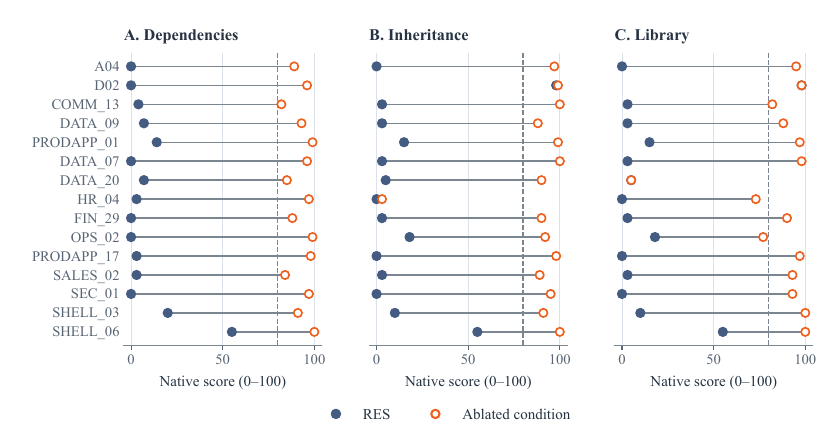}
\caption{\textbf{Per-task effects of the component interventions.} Filled blue markers show the corresponding full RES control; hollow orange markers show the ablation. Segments connect the same task, and dashed lines mark the 80-point success threshold. All panels use the task order in Table~\ref{tab:per-task-scores}.}
\label{fig:per-task-ablations}
\end{figure}

Table~\ref{tab:per-task-scores} reports native scores for all 15 selected tasks. Figure~\ref{fig:per-task-ablations} pairs each intervention with its corresponding RES control, keeping the task order fixed across comparisons.

\begin{table}[tb!]
\centering\small
\caption{\textbf{Per-task native scores for the 15-task component ablations.} DeepSeek-V4.1-Flash receives one 50-turn attempt per environment; scores use a 0--100 scale, with success at 80 or above. Original is the unmodified task. Saved and Rebuilt denote full RES controls; w/o deps. removes dependencies from Saved, w/o rec. independently enhances Original and evaluates the predeclared last construction, and Basic rebuilds with six operators. The rebuilt variants share planned depth $d_i$.}
\label{tab:per-task-scores}
\begin{tabular*}{\linewidth}{@{\extracolsep{\fill}}lrrrrrrr@{}}
\toprule
Task & $d_i$ & Original & Saved & w/o deps. & Rebuilt & w/o rec. & Basic \\
\midrule
A04 & 10 & 93.00 & 0.00 & 89.00 & 0.00 & 97.00 & 95.00 \\
D02 & 4 & 96.00 & 0.00 & 96.00 & 98.00 & 99.00 & 98.00 \\
COMM\_13 & 5 & 100.00 & 4.00 & 82.00 & 3.00 & 100.00 & 82.00 \\
DATA\_09 & 8 & 86.00 & 7.00 & 93.00 & 3.00 & 88.00 & 88.00 \\
PRODAPP\_01 & 3 & 92.00 & 14.00 & 99.00 & 15.00 & 99.00 & 97.00 \\
DATA\_07 & 3 & 96.00 & 0.00 & 96.00 & 3.00 & 100.00 & 98.00 \\
DATA\_20 & 4 & 84.00 & 7.00 & 85.00 & 5.00 & 90.00 & 5.00 \\
HR\_04 & 4 & 98.00 & 3.00 & 97.00 & 0.00 & 3.00 & 73.00 \\
FIN\_29 & 3 & 100.00 & 0.00 & 88.00 & 3.00 & 90.00 & 90.00 \\
OPS\_02 & 4 & 92.00 & 0.00 & 99.00 & 18.00 & 92.00 & 77.00 \\
PRODAPP\_17 & 3 & 99.00 & 3.00 & 98.00 & 0.00 & 98.00 & 97.00 \\
SALES\_02 & 4 & 93.00 & 3.00 & 84.00 & 3.00 & 89.00 & 93.00 \\
SEC\_01 & 4 & 98.00 & 0.00 & 97.00 & 0.00 & 95.00 & 93.00 \\
SHELL\_03 & 4 & 100.00 & 20.00 & 91.00 & 10.00 & 91.00 & 100.00 \\
SHELL\_06 & 3 & 100.00 & 55.00 & 100.00 & 55.00 & 100.00 & 100.00 \\
\bottomrule
\end{tabular*}

\end{table}

\subsection{Execution Cost and Score Sensitivity}
\label{app:ablation-cost}

\paragraph{Token and interaction cost.}
Table~\ref{tab:ablation-cost} aggregates input and output tokens across each target rollout, including cached input and reasoning tokens once in the reported totals. Relative token cost is the ratio of task means. A model turn may contain several tool calls, and these measurements include retries and unfinished trajectories. Every saved RES task reaches the 50-turn limit; removing its dependencies reduces the mean to 19.1 turns. The component comparisons comprise 90 rollouts.

\begin{table}[tb!]
\centering\small
\setlength{\tabcolsep}{4pt}
\renewcommand{\arraystretch}{1.08}
\caption{\textbf{Target execution cost for the 15-task component comparisons.} Values are task means for DeepSeek-V4.1-Flash with one 50-turn attempt per environment, including failed and turn-limited attempts. Saved RES is the dependency-removal control; Rebuilt RES is shared by the independent-construction and basic-library comparisons. Tokens sum target input and output; k denotes thousands. Relative tokens divides the condition's mean by the Original mean before rounding. Proposer construction and grading costs are excluded.}
\label{tab:ablation-cost}
\begin{tabular*}{\linewidth}{@{\extracolsep{\fill}}lrrrr@{}}
\toprule
Condition & Turns & Tool calls & Tokens (k) & Relative tokens ($\times$) \\
\midrule
Original & 2.87 & 6.67 & 8.43 & 1.00 \\
Saved RES & 50.00 & 226.93 & 3115.42 & 369.40 \\
w/o dependencies & 19.07 & 85.60 & 723.76 & 85.82 \\
Rebuilt RES & 49.73 & 163.93 & 1644.20 & 194.96 \\
w/o inheritance & 21.20 & 90.40 & 651.09 & 77.20 \\
Basic library & 18.53 & 39.60 & 227.01 & 26.92 \\
\bottomrule
\end{tabular*}

\end{table}
\FloatBarrier

\begin{table}[H]
\centering\footnotesize
\setlength{\tabcolsep}{3pt}
\renewcommand{\arraystretch}{0.94}
\caption{\textbf{Success counts under alternative native-score thresholds on Claw-Eval-Live.} Each entry counts tasks with score $\geq\tau$ over the cohort. Component rows use 15 paired tasks; depth rows use 105 reevaluated seeds with stopped environments retained. Counts use saved scores.}
\label{tab:threshold-sensitivity}
\begin{tabular*}{\linewidth}{@{\extracolsep{\fill}}lrrr@{}}
\toprule
Condition & $\tau=0.70$ & $\tau=0.80$ & $\tau=0.90$ \\
\midrule
Original & 15/15 & 15/15 & 13/15 \\
Saved RES & 0/15 & 0/15 & 0/15 \\
w/o dependencies & 15/15 & 15/15 & 10/15 \\
Rebuilt RES & 1/15 & 1/15 & 1/15 \\
w/o inheritance & 14/15 & 14/15 & 12/15 \\
Basic library & 14/15 & 12/15 & 10/15 \\
\midrule
Depth cap 0 & 92/105 & 87/105 & 68/105 \\
Depth cap 1 & 68/105 & 63/105 & 45/105 \\
Depth cap 2 & 18/105 & 18/105 & 15/105 \\
Depth cap 3 & 8/105 & 8/105 & 4/105 \\
Depth cap 5 & 4/105 & 4/105 & 1/105 \\
Depth cap 10 & 6/105 & 5/105 & 2/105 \\
\bottomrule
\end{tabular*}

\end{table}

\end{document}